\documentclass{article}

\usepackage{arxiv}

\usepackage[utf8]{inputenc} 
\usepackage[T1]{fontenc}    
\usepackage{hyperref}       
\usepackage{url}            
\usepackage{booktabs}       
\usepackage{amsfonts}       
\usepackage{nicefrac}       
\usepackage{microtype}      
\usepackage{lipsum}
\usepackage[normalem]{ulem}
\usepackage{graphicx}

\graphicspath{ {./images/} }

\title{Convolutional Lie Operator for Sentence Classification}

\author{
    Daniela N. Rim \\
  School of Computer Science and Electrical Engineering\\
  Handong Global University \\
  Pohang, Republic of Korea \\
  \texttt{danielarim@handong.ac.kr} \\
   \And
    Heeyoul Choi \\
  School of Computer Science and Electrical Engineering\\
  Handong Global University \\
  Pohang, Republic of Korea \\
  \texttt{heeyoul@gmail.com} \\
}

\begin{document}
\maketitle
\begin{abstract}
Traditional Convolutional Neural Networks have been successful in capturing local, position-invariant features in text, but their capacity to model complex transformation within language can be further explored.
In this work, we explore a novel approach by integrating Lie Convolutions into Convolutional-based sentence classifiers, inspired by the ability of Lie group operations to capture complex, non-Euclidean symmetries. Our proposed models SCLie and DPCLie empirically outperform traditional Convolutional-based sentence classifiers, suggesting that Lie-based models relatively improve the accuracy by capturing transformations not commonly associated with language. Our findings motivate more exploration of new paradigms in language modeling.
\end{abstract}


\section{Introduction}
Since the introduction of the word-to-vector or \textit{Word2Vec} representations \cite{mikolov2013efficient}, models could train high-dimensional vectors in a one-to-one relation with word units (a.k.a. language tokens.) The result is vector representations that retain some of the word properties, such as synonymity and antonymyty (correlated with cosine similarities of the vectors.) Analogously, for sentence classification tasks, it is desirable to have a vector representation of a whole sentence that retains its features and can be contrasted/compared with other sentences' representations.

In a sentence classification setting, linearly combining meaningful token vectors does not result in a synergy, that is, the meaning of a sentence in the representation space is not that of the aggregation of its parts. Thus, NLP models such as Recurrent Neural Networks \cite{socher2013recursive}, Convolutional Networks \cite{Kim2014ConvolutionalNN}, and more recently, Transformer-based \cite{vaswani2017attention} models such as Sentence BERT (SBERT), \cite{reimers2019sentence} have been used on top of this space to extract the sentence information between tokens.

While current NLP models perform well on sentence classification, there remain challenges in efficiently handling the complex structures and invariant properties inherent to natural language. Furthermore, NLP models can always benefit from innovative techniques and ideas that enhance their generalization and robustness. Additionally, nuances in language could be viewed as transformations that are often neither linear nor Euclidean in nature, which is not yet accommodated by the current NLP models.

To address these challenges, we explore an alternative approach to sentence classification via Convolutional Networks by introducing Lie Convolutions to the architecture. Lie Convolutions leverage Lie group structures, allowing the model to capture non-Euclidean symmetries, which we hypothesize may be beneficial for representing complex linguistic patterns. Lie groups, which encompass transformations like rotations and translations, provide a mathematical framework that could lead to smoother, more transformation-robust representations in sentence classification tasks.

Our empirical results show that introducing Lie Convolutions to represent and classify sentences improves performance over traditional Convolutional layers approaches, suggesting that Lie operators can capture sentence-level features that enhance sentence embeddings. We propose that this framework Convolutional-Lie (CLie) and its deeper variant DPCLie, offer a novel, more expressive, and robust means of modeling sentence representations compared to conventional convolutional layers.

The contributions of this paper are as follows: \begin{itemize} 
\item We introduce Lie Convolutions to sentence classification tasks, demonstrating how this approach can improve the robustness and expressiveness of Convolutional-based architectures. 
\item We empirically validate the effectiveness of our proposed models on multiple benchmark sentence classification datasets, showing improved accuracy and smoother representations in comparison to baseline models. 
\item We investigate the potential of Lie Convolutions for capturing symmetry properties in language data, which could be a starting point for considering non-Euclidean representations when dealing with language. 
\end{itemize}

The paper is organized as follows: Section \ref{dnnSC} covers related works, Section \ref{meths} outlines our methodology, explaining our proposed SCLie and DPCLie architectures. In Section 4, we present our experiments and Section 5 provides conclusions and directions for future work.

\section{Related Works} \label{dnnSC}

ConvNets consist of convolutional layers that learn data features via filter (or kernel) optimization. These kernels convolve along input features and provide translation-equivariant responses (feature maps.) Translation-equivariance means that when the input is shifted, the resulting feature maps shift in the same manner, allowing ConvNets to detect features regardless of their position \cite{lecun1998gradient,lecun2015deep}. 

The standard convolution operation in ConvNets is a special case of a broader concept known as group\footnote{\textit{group} as defined in the field of group theory} convolutions. Group convolutions are general linear transformations that remain equivariant to a specific group of transformations, and can allow neural networks to recognize more complex symmetries beyond simple translations. This concept is effectively utilized in Group Equivariant Convolutional Networks (G-CNNs) \cite{cohen2016group}, which extend the capabilities of traditional ConvNets by capturing patterns under various transformations, enhancing the model’s ability to generalize across different symmetries.

With this motivation, Finzi et al. (2020) proposed a new convolutional layer equivariant to transformations from Lie groups. Lie groups are mathematical structures that combine algebraic and topological properties: the properties of both a group (with a group operation) and a smooth manifold. That is, these are groups that are also differentiable manifolds, meaning that the group operations (multiplication and inversion) are smooth. Examples of Lie groups include the set of all rotations in 3D space (SO(3)), the set of all translations and rotations (SE(3)), etc.

When dealing with functions defined on Lie groups, traditional convolution does not apply directly because the underlying space is not a Euclidean space but rather a manifold with its own geometric structure. Lie convolution adapts the convolution operation to this context by integrating a function over the group, using the group's natural measure (Haar measure), and respecting the group's structure \cite{lu2020survey}.

\subsection{Convolutional Networks in Sentence Classification}
ConvNets have also proven effective in Natural Language Processing (NLP) tasks, despite their most common use in Computer Vision tasks. Text data, like visual data, contains local and position-invariant patterns, making ConvNet a popular architecture for text classification \cite{survey2021}. One of the earliest and simplest ConvNet models for text classification was proposed in \cite{Kim2014ConvolutionalNN}, where a word embedding matrix was constructed from a sentence using pre-trained unsupervised vectors from Word2Vec, followed by a convolutional layer on top and a max-pooling layer to generate feature maps capable of capturing relations of words within the sentence. Subsequent ConvNet architectures have made architectural improvements by adding a dynamic pooling scheme, adding bottlenecks for efficiency, using other word embedding schemes \cite{liu2017deep, johnson2014effective, johnson2017deep}, and Character-based models \cite{zhang2015character, kim2016character}.

More than 150 model architectures for text classification have been proposed containing and/or combining Feed-forward networks, Recurrent Neural Networks, Graph Neural Networks, Convolutional Neural Networks (ConvNets or CNNs), and Transformer-based models. For a comprehensive survey, please refer to \cite{survey2021}. In this work, we will focus on the ConvNet-based models, specifically exploring the integration of convolutional Lie layers \cite{finzi2020generalizing} within a sentence classifier. This approach aims to benefit from smoother and more robust-to-transformation representations, potentially revealing symmetries that are not readily apparent in language data in the context of convolutional networks. Although Transformer-based models hold the state-of-the-art performance in sentence classification, our goal is to introduce the Lie convolutional layer within a familiar architecture to evaluate its specific contributions.


\section{Method}

\subsection{Lie Convolutions} \label{meths}

A Lie group $G$ is composed of symmetry transformation elements $g\in G$ forming a smooth manifold (continuous and differentiable structure.) The elements of $G$ are not necessarily a vector space, so to operate within elements of the group we introduce the tangent space at the identity of the Lie group $G$, that is, the Lie algebra $\mathfrak{g}=T_{id}G$. An exponential map $\exp:\mathfrak{g}\rightarrow G$ and its inverse logarithmic mapping $\log:G \rightarrow \mathfrak{g}$ can be defined to operate and move between the tangent space $\mathfrak{g}$ and $G$. 

Adopting the definition of group convolution between two functions $k,g: G \rightarrow \mathbb{R}$ of \cite{finzi2020generalizing}, we formally define the group convolution $C$ of $k$ with $g$ at any $h$ in $G$ given by:
\begin{equation}
    C(h) =(k\ast g)(h) = \int_{G} k(v^{-1}h)g(v)\,d\mu (v),
\label{eq:integrl}
\end{equation}
where $\mu$ is the Haar measure on the group $G$ and $v^{-1}h$ represents the group operations (typically, inverting $v$ and applying it to $h$). This operation is analogous to the standard convolution but generalized to any transformation in $G$. 

Suppose we have data in the form of a set of input pairs $(x_i, f_i)^n_{i=1}$ where $x_i\in \chi$ are spatial coordinates and $f_i \in \mathcal{F}$ are feature values. For example, if we are dealing with text data, $x_i$ could be a token with a feature vector associated with it. The data can be described as a single feature map $\mathcal{F}_\chi:x_i\rightarrow f_i$ (for example, a token in a sentence mapped to a meaningful embedding vector $f_i$, subject to invariant transformations like rephrasings). If $\chi$ is a homogeneous space of $G$, we can define a lift function $\mathrm{L}(x)={\{h \in G:h_o=x\}}$, that is, all the transformations $h$ in $G$ that map the origin to $x$. This enables the mapping between sets of inputs and transformations in $G$: $\{(x_i,f_i)\}^N_{i=1}\rightarrow \{(h_{ik},f_i)\}^{N,K}_{i=1, k=1}$, where $K$ is the number of group elements. Simply put, by defining $L(x)$, we set an origin in $G$ with respect to which we can describe any input $x$ in terms of transformations $h$ in $G$. Following the text application, this could be useful for the model to capture transformation-invariant representations of text data.

In order to make the integral in Eq. \ref{eq:integrl} feasible, \cite{finzi2020generalizing} proposed to model the kernel $k$ by mapping onto the Lie Algebra $\mathfrak{g}$ by restricting the approach to Lie groups with surjective exponential maps. Furthermore, the integral is defined over a small neighborhood ($nbh$) of $v$ instead of the whole $G$, which is as follows: 
\begin{equation}
  C_{lie}(h) = (k_\theta \ast g)(h) = \int_{nbh(v)} k_\theta(v^{-1}h)g(v)\,d\mu (v),
\label{eq:lie}
\end{equation}
with $k_\theta (h) = (k\odot \exp)_\theta (\log h)$, and $(k\odot \exp)_\theta$ parameterized by an MLP. The integral is discretized via Monte Carlo to enable its computation. The proof of the validity of each approach is explained and shown in detail in \cite{finzi2020generalizing}.

\subsection{Proposed method}

The discrete nature of language has been a challenge for NLP methods since its introduction. The most accepted approach to enable training via direct backpropagation is the usage of continuous vectors for word (token) representations, which can be extended to sentence representations. Like Word2Vec methods, it is assumed that language nuances are represented in this continuous data space. We hypothesize that language representations not only adapt well to these continuous embedding spaces but may also benefit from a smooth manifold assumption, as enabled by Lie groups. By operating in a structured, continuous space, Lie groups could capture underlying symmetries and transformations in language data, potentially leading to similar or even improved performance over conventional methods in tasks that involve nuanced relational patterns and invariant features.

\begin{figure}[h]
  \centering
  \includegraphics[width=0.9\linewidth]{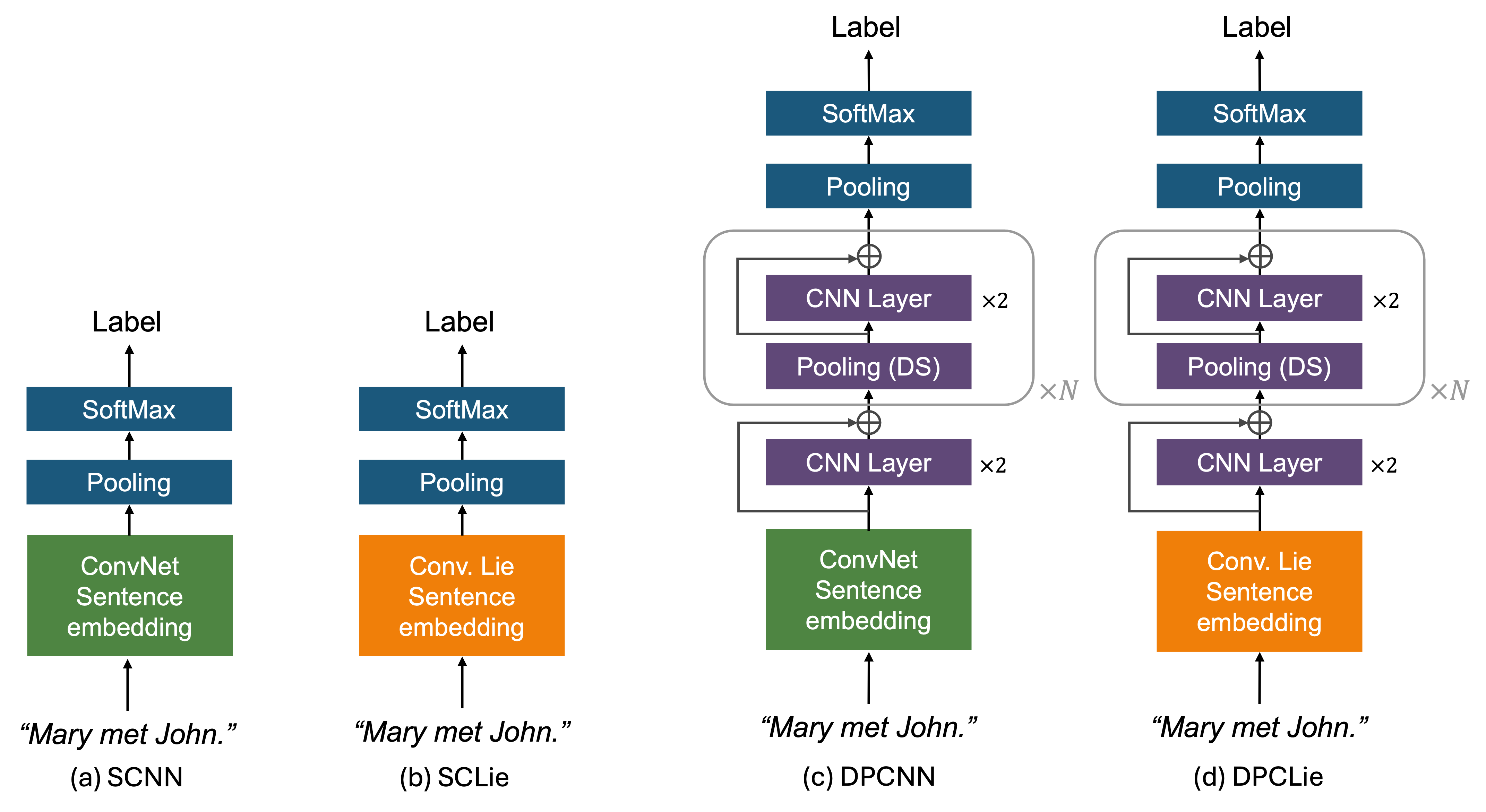}
  \caption{We show the conventional convolutional-based sentence classifications \textbf{(a)} SCNN and \textbf{(c)} DPCNN next to our proposed models incorporating a Convolutional Lie layer. \textbf{(b)} \textbf{SCLie}, a one-layer Lie convolutional layer sentence classifier, convolves the input embeddings with multiple filter widths and feature maps to obtain sentence representations and finally makes a classification after a pooling and a softmax layer. \textbf{(d)} \textbf{DPCLie}, a deeper version of \textbf{(b)} which adds a block of Convolutional Layers that downsample (DS) the representations, to add depth to the architecture without significantly increasing the computational burden.}
\label{fig:model_arch}
\end{figure} 

Figure \ref{fig:model_arch} illustrates the general architecture of our proposed models. A sentence is given as input to the models, and the concatenated words in each sentence from a batch are processed as a grid via a convolutional (conventional or Lie) layer, which results in feature vectors (from different kernels). These feature vectors are max-pooled and concatenated into one sentence representation vector. We could directly obtain a final output classification by a fully connected layer with softmax on the feature matrix (Figure \ref{fig:model_arch} (a, b)) or adopt the Deep Pyramid Convolutional Neural Networks (DPCNN) architecture \cite{johnson2014effective}. The latter forwards the feature map through multiple convolutional layers, iteratively downsampling through max-pooling after each convolutional block, with residual connections ensuring the preservation of essential features across layers, before flattening and passing it to a fully connected layer and softmax as in Figure \ref{fig:model_arch} (c, d). 

Our two proposed architectures involve a Lie convolutional layer instead of a conventional convolutional layer. We call these two proposed architectures Convolutional-Lie (CLie) and Deep Pyramidal Convolutional-Lie (DPCLie) as in Figure \ref{fig:model_arch}.

The convolutional Lie layer in Figure \ref{fig:model_arch} (b,d) works as follows. Like other classification models, our input is a sequence of word vectors $x_1, x_2, ..., x_n \in \mathbb{R}^d$. However, we introduce a set of transformations $g_1,g_2,... g_n$ from a group $G$ to each representation.

Following the kernel $k_\theta$ definition given in Eq. \ref{eq:lie}, a feature $c_i$ is generated from a window of words $x_{i:i+l-1}$ and the corresponding group elements $g_{i:i+l-1}$ by:
\begin{equation}
    c_i = f\big( \sum^{l-1}_{j=0} k_\theta (g_{i+j})\cdot x_{i+j} + b\big),
\label{eq:proposed}
\end{equation}
where $f$ is a non-linear activation function, $k_\theta (g_{i+j})$ is the filter applied at position $j$, transformed by the Lie group element $g_{i+j}$, $l$ is the dimension of the kernel, and $b$ is a bias term.

\begin{figure}[h]
  \centering
  \includegraphics[width=0.95\linewidth]{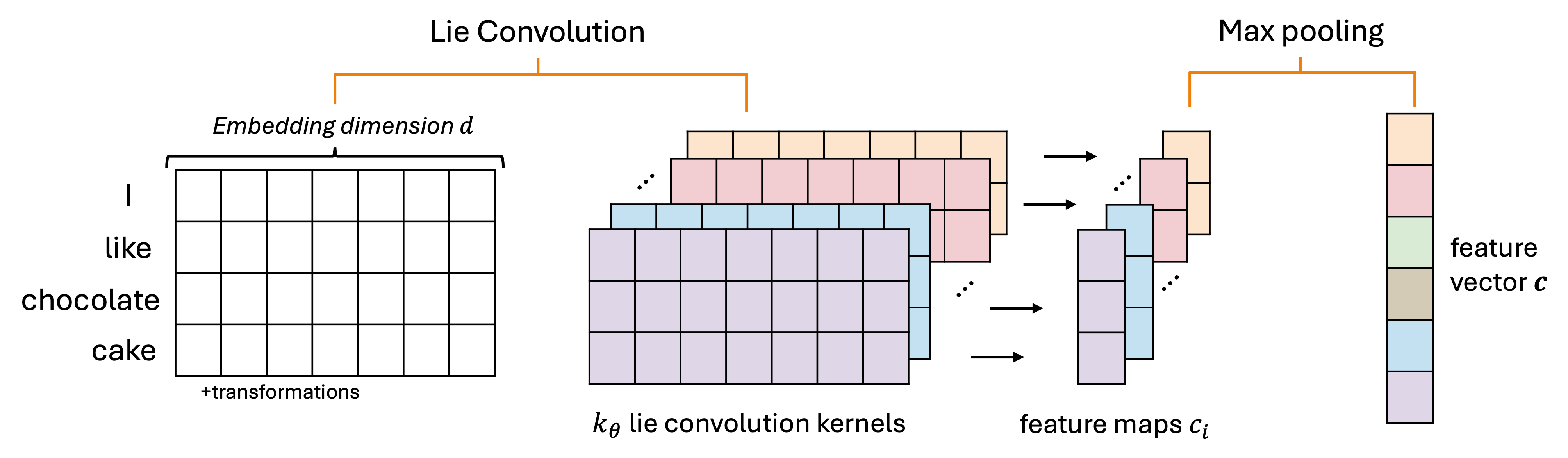}
  \caption{Proposed Lie convolution layer for sentence classification. We show a simplified example for one sentence in the batch, which is forwarded through the embedding layer to obtain a grid-type tensor. The embedding layer contains a data transformation analogous to a Lie group. The grid is convolved with $k_\theta$ dynamic filters/kernels of different sizes to capture different features. These representations are max-pooled to obtain a vector representation of the whole text, finally obtaining a classification.}
\label{fig:sent_rep}
\end{figure}

Then, the operation continues as the original ConvNet, with the filter applied to all possible windows in the sentence. The result is a feature map 
\begin{equation}
    \mathbf{c} = [c_1,c_2,...,c_{n-l+1}],
\end{equation}
with $\mathbf{c}\in \mathbb{R}^{n-l+1}$. This process is shown for only one sentence in Figure \ref{fig:sent_rep}. The subsequent computations remain the same, with a max-pooling operation $\hat{\mathbf{c}}= \max\{\mathbf{c}\}$. The use of multiple kernels results in a feature matrix
\begin{equation}
    \mathbf{C}=[\hat{\mathbf{c}}_1, \hat{\mathbf{c}}_2,..., \hat{\mathbf{c}}_s],
\end{equation}
with $s$ different filters.

\section{Experiments}
In order to test the performance of our approach with SCNN \cite{Kim2014ConvolutionalNN} and DPCNN \cite{johnson2017deep}, we use seven evaluation datasets commonly used in sentence embedding evaluation toolkits such as SentEval\cite{conneau2018senteval}:
 
 \begin{itemize}
     \item \textbf{Customer Reviews (CR):} binary classification (positive/negative) of product reviews \cite{hu2004}.
    \item \textbf{Multi-Perspective Question Answering (MPQA):} binary classification of Opinion polarity detection subtask of the MPQA dataset \cite{dolan2004unsupervised}.
    \item \textbf{Text REtrieval Conference (TREC):} classification of questions into 6 categories- \textit{abbreviation, entity, description and abstract concept, human description, location, and numeric value }\cite{liroth2002}.
    \item \textbf{Movie Reviews (MR):} binary classification (positive/negative) of one-sentence movie reviews \cite{panglee2025}
    \item \textbf{Stanford Sentiment Treebank binary (SSTb):} binary classification of movie reviews \cite{socher2013recursive}. 
    \item \textbf{Subjectivity (Subj):} binary classification of objective/subjective sentences \cite{pang2004sentimental}.
\end{itemize}

The sizes of the dataset and their vocabulary are shown in Table \ref{tab:voc_size}. The token representations were initialized with the pre-trained word2vec vectors from the bag-of-words (CBOW) model \cite{mikolov2013efficient} trained with the Google News dataset (around 100 billion words.) \footnote{\url{https://code.google.com/archive/p/word2vec/}}. The vocabulary words not present in the pre-trained database were initialized randomly.

\begin{table*}[h]
\centering
\begin{tabular}{ccccccc}
\hline
\textbf{Dataset} & \textbf{CR} & \textbf{MPQA} & \textbf{TREC} & \textbf{SSTb}  &\textbf{MR} &\textbf{Subj}  \\ \hline
\textbf{Number of Sentences}   & 3775    & 10606          & 5952    &9613  & 10662& 10000  \\ 
\textbf{Voc. Size}   & 5057        & 5195          & 9330     &9613  & 16758& 18999  \\ 
\hline
\end{tabular}
\caption{Datasets used for the evaluation of our proposed model. We show the total number of sentences and the vocabulary size of each datastet.}
\label{tab:voc_size}
\end{table*} 

In the implementation of the Lie convolutional sentence embedding layer for SCLie and DPCLie, ReLU was used as the activation function, with filter windows of sizes 3, 4, and 5, each generating 100 feature maps, similar to the convolutional layer sentence embedding setup in SCNN and DPCNN. For SCNN and our proposed SCLie, the dropout rate was 0.5, and the mini-batch size was 50. The training was conducted using Adadelta, with early stopping to mitigate the increased parameter effect of the Lie model. All words, including unknown ones initialized randomly with word2vec, are fine-tuned for each task. Additionally to the convolution models (conventional and Lie), we ran a \textit{Linear model} version that replaced the Convolution layer with a Fully connected Neural Network (with the same number of parameters) to ensure the ConvNet contribution to the sentence classification.

For the DPCNN and DPCLie, we use a mini-batch stochastic gradient descent (SGD) with a momentum 0.9 and early stopping. The learning rate was kept constant for the first 45\% of the training epochs and then reduced to 10\% for the remainder. The mini-batch size was set to 100 for DPCNN and 64 for DPCLie. Regularization was applied using weight decay with a parameter of 0.0001 and dropout with a rate of 0.5, applied to the input of the top layer. The output dimensionality of the ConvNet layers was kept at 250, and the number of convolutional blocks ($N$ in Figure \ref{fig:model_arch} (c,d)) was set to $N=1$.

The architectures were trained with GPUs \textit{NVIDIA GeForce GTX 1080 Ti} and \textit{NVIDIA TITAN Xp}. The code implementation was done using Pytorch, and the code is available in git\footnote{\url{}}. 

\subsection{On Symmetry in Representations}

As mentioned in the introduction section, a key advantage of Lie operators is their ability to capture symmetries in representation space. We hypothesized that Lie operators can hint at non-Euclidean symmetry properties in language. To investigate this, we assess whether the trained DPCLie model can generate sentence representations that reflect symmetrical properties at a sentence level. We use a curated version of the Symmetry Inference Sentence (SIS) dataset \cite{tanchip2020inferring}, originally consisting of 400 sentence pairs with symmetrical/asymmetrical verbs, and a Likert symmetry score between 1 and 5 (1 indicating identical meaning, and 5 indicating completely different meaning.) The sentences have a general format of $[entity_1] [verb] [entity_2]$ and a syntactically alternated version $[entity_2] [verb] [entity_1]$. Examples from the dataset are shown in Table \ref{tab:examples}.

\begin{table}[]
\begin{tabular}{ll|c}
\hline
\textbf{Sentence 1} & \textbf{Sentence 2} & \textbf{Score} \\ \hline
\begin{tabular}[c]{@{}l@{}}The children and their parents \textbf{love} \\  each other.\end{tabular} &
  \begin{tabular}[c]{@{}l@{}}The parents and their children \textbf{love} \\  each other.\end{tabular} & 1 \\
\begin{tabular}[c]{@{}l@{}}Mark \textbf{hit} me two or three times in the\\  same period.\end{tabular} &
  \begin{tabular}[c]{@{}l@{}}I \textbf{hit} Mark two or three times in the\\ same period.\end{tabular} &  5 \\
  \hline
\end{tabular} \caption{Examples of the SIS dataset.}
\label{tab:examples}
\end{table}

For this study, we filtered out all the intermediate scores (2-4) from the SIS dataset and sampled 200 sentence pairs with scores of 1 and 5.

\section{Results}

\subsection{Sentence Classification}
We measure the accuracy of the sentence classifications as shown in Table \ref{tab:results}. All the models were implemented from scratch.

\begin{table*}[h]
\centering
\begin{tabular}{ccccccc|c}
\hline
& \textbf{CR} & \textbf{MPQA} & \textbf{TREC} & \textbf{SSTb}  & \textbf{MR}  & \textbf{Subj} & \textbf{AVG}.\\ \hline
\textbf{Linear} & 0.818 & 0.863 & 0.868 & 0.831 & 0.768 & 0.907 & 0.843\\
\textbf{SCNN } & 0.815 & 0.829 & \uline{0.932} & 0.835 & 0.797  &0.930 & 0.856\\
\textbf{SCLie } & \uline{0.836} & \uline{0.872}  & 0.922  & \uline{0.843} & \uline{0.798} & \uline{0.931} & \textbf{0.867}\\
\hline
\textbf{DPCNN} & 0.824 &0.850 & 0.902 & 0.821  & 0.772 & 0.924 & 0.849\\
\textbf{DPCLie} & \uline{0.826}& \uline{0.871}& \uline{0.932} & \uline{0.851} & \uline{0.774} & \uline{0.925} & \textbf{0.863}\\
\hline
\textbf{SBERT} & \textit{0.900} & \textit{0.903} & \textit{0.874} &\textit{ 0.907} & \textit{0.849} & \textit{0.945} & \textit{0.896}	\\
\hline
\end{tabular}
\caption{Accuracy for sentence classification on six common datasets for classification tasks. We compare the proposed models SCLie and DPCLie with their baselines and show the average score of all tasks (\textbf{AVG}.)}
\label{tab:results}
\end{table*}

As shown in the Table \ref{tab:results}, there is a general increase in accuracy across most datasets. Furthermore, when comparing \textbf{SCNN} with \textbf{DPCNN}, there is not a significant improvement by adding depth to the architecture.

It is important to note that our proposed models introduced more trainable parameters due to dynamically trained kernels of the Lie operator (Eq. \ref{eq:proposed}). The number of parameters varies depending on the vocabulary size of each dataset, but on average the proposed method of Lie convolutional layers doubled the number of parameters. Our models were set with the same hyperparameters as the original models as in \cite{Kim2014ConvolutionalNN} and \cite{johnson2017deep}. To examine whether the better performance of our proposed model was solely due to the increased number of parameters, we doubled the number of channels in DPCNN to match the number of trainable parameters in DPCNN with our implementation DPCLie. When trained with double parameters, DPCNN had an average accuracy score of $0.841$, which is marginally lower than what was shown in Table \ref{tab:results}. This result says that the improvement by CLie is not simply based on more parameters. 

We also show the performance of the Transformer-based small SBERT model for reference reasons. The number of parameters is $18\times$ larger than the CLie and $9\times$ larger than DPCLie.

\begin{figure}[h]
  \centering
  \includegraphics[width=0.7\linewidth]{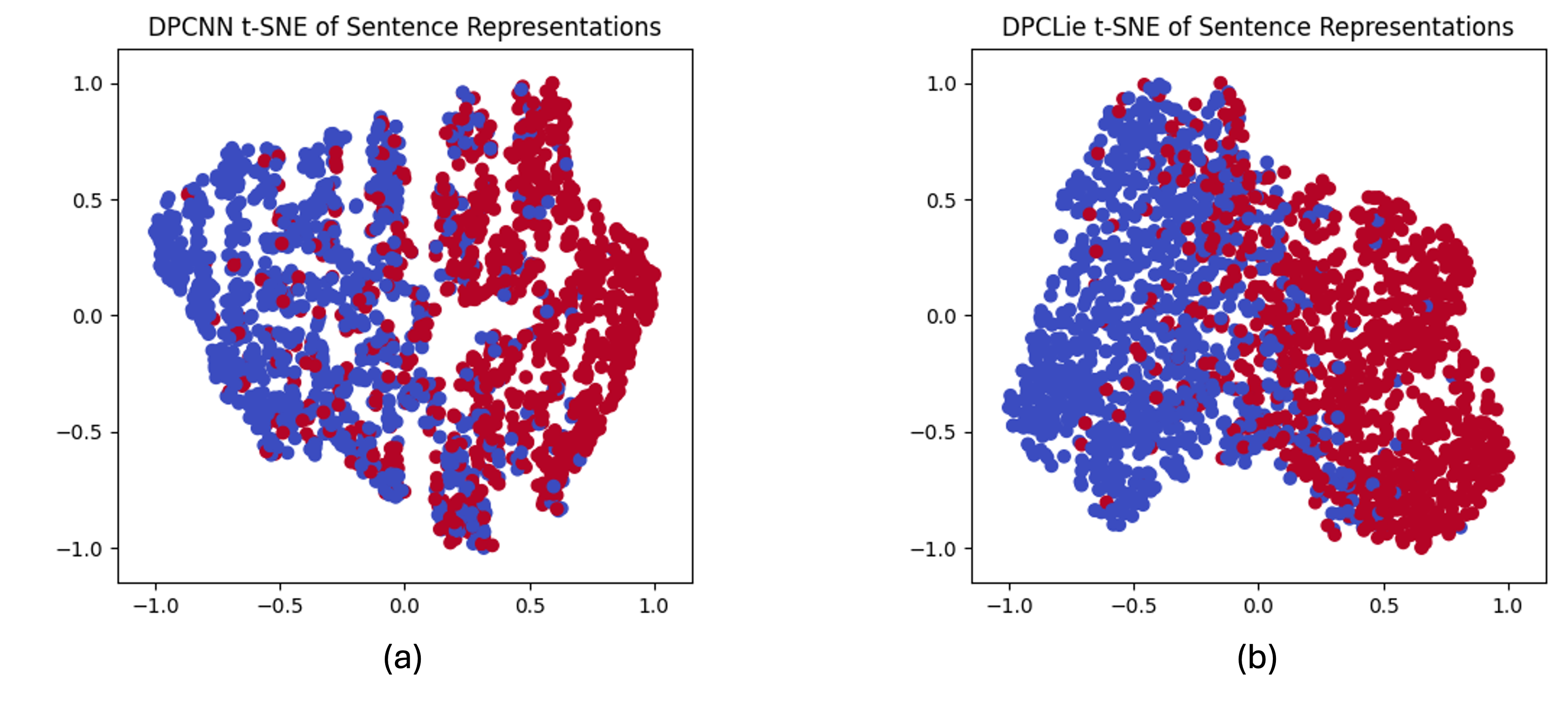}
  \caption{Visualizations of 2-D t-SNE applied to the sentence representations of the trained DPCNN (a) and DPCLie (b) on the SST dataset. The data points are colored by the binary classification labels corresponding to each sentence.}
\label{fig:smooth}
\end{figure}

As explained in Section \ref{dnnSC}, one of the assumptions when working with Lie algebra is that the data lies on a smooth manifold. After training our models with DPCNN and DPCLie, we compared the trained representations made by both models on the test sentences prior to classification. We applied the dimensionality reduction technique t-SNE to visualize these representations (after normalization), as shown in Figure \ref{fig:smooth}. The sentence representations produced by DPCNN appear more clustered, with visible regions of discontinuity, whereas DPCLie exhibits smoother transitions and more continuous structures.

\subsection{Symmetry}
We obtained the sentence representations of 200 sentence pairs from the SIS dataset using the trained DPCNN and DPCLie architectures. Each pair contains one sentence and its syntactically alternated version, as shown in Table \ref{tab:examples}. We computed the cosine similarity for every pair of representations
and obtained a similarity score. We then compared these scores with the ground truth labels and measured the Spearman and Pearson correlations (Table \ref{tab:sis_table}.)

\begin{table}[h]
\centering
\begin{tabular}{lcc}
\hline
                          & \textbf{DPCNN} & \textbf{DPCLie} \\ \hline
\textbf{Pearson Correlation }      & 0.23  & 0.26   \\
\textbf{Spearman Rank Correlation} & 0.24  & 0.24  \\
\hline
\end{tabular}
\caption{Pearson Correlation and Spearman Rank Correlation of the trained DPCNN and DPCLie using part of the SIS dataset. For each pair of sentences, we obtain the cosine similarity of their sentence representations, and correlate these scores with the ground truth symmetry labels.}
\label{tab:sis_table}
\end{table}

DPCLie has a higher Pearson correlation, indicating a slightly stronger linear relationship between the cosine similarity of its representations and the ground truth labels. Both models performed similarly in capturing monotonic relationships in the data, as shown by the Spearman Rank correlation.

While these scores do not overall indicate a strong relation between representation similarity and symmetry scores, it is worth noting neither model was explicitly trained to recognize symmetry. However, DPCLie demonstrates a relative improvement in capturing the difference between symmetrically related sentences, which may provide an advantage in capturing nuanced relationships in language.

\section{Conclusion}
In this work, we introduced a novel approach to sentence classification by incorporating Lie convolutions into ConvNet-based architectures in order to capture complex, non-Euclidean relations in language data. Based on Lie-group theory, our proposed models SCLie and DPCLie generate sentence representations that demonstrate smoother transitions in the representation space. Experiments in multiple benchmark datasets show that our models outperform the original ConvNet architectures.

Furthermore, our exploration of symmetry properties in sentence representations suggests that Lie-based convolutions can better capture nuanced relationships, even in the absence of explicit training for symmetry recognition. Although modest, this highlights a potential of non-Euclidean transformations in enhancing model expressiveness and robustness in language processing. 

Future work could extend this approach by further focusing Lie convolution layers to capture specific symmetrical transformations via a symmetry loss, and comparing their representations to those obtained by contextualized large transformer-based models. Furthermore, the application of Lie operators could be explored in other natural language areas such as language modeling. 
Overall, our findings motivate deeper exploration of non-Euclidean frameworks in language processing, potentially leading to new representations and training paradigms.

\section{Acknowledgments}
This research was supported by Basic Science Research Program through the National Research Foundation of Korea funded by the Ministry of Education (NRF-2022R1A2C1012633), and the MSIT(Ministry of Science, ICT), Korea, under the Global Research Support Program in the Digital Field program(RS-2024-00431394) supervised by the IITP(Institute for Information \& Communications Technology Planning \& Evaluation).

\bibliographystyle{ACM-Reference-Format}
\bibliography{custom.bib}

\end{document}